Methods for Mapping Forest Disturbance and Degradation from Optical Earth Observation Data: a Review

**Dr. Manuela Hirschmugl[a, *], DI Heinz Gallaun[a], Dr. Matthias Dees[b], Dr. Pawan Datta[b], Mag. Janik Deutscher[a], Dr. Nikos Koutsias[c], Prof. Dr. Mathias Schardt[a]**

[*] corresponding author

[a] JOANNEUM RESEARCH, Steyrergasse 17, 8010 Graz.

Email: (manuela.hirschmugl, heinz.gallaun, mathias.schardt, janik.deutscher)@joanneum.at

Phone: 0043-316-876-(1707, 1757, 1754, 1776); Fax: 0043-316-876-91720

[b] Albert-Ludwigs-University Freiburg, Chair of Remote Sensing and Forest information Systems, Tennenbacher Str. 4 79085 Freiburg, Germany.....

Email: (matthias.dees, pawanjeet.datta)@felis.uni-freiburg.de

Phone: 0049-761-203-(3697, 3700); Fax: 0049-761-203-3701

[c] University of Patras, Department of Environmental and Natural Resources Management, G. Seferi 2, Agrinio, GR-30100, Greece

Email: nkoutsia@upatras.gr

Phone: 0030-26410-74201; Fax: 0030-26410-74176

**Abstract**:

Purpose of review: This paper presents a review of the current state of the art in remote sensing based monitoring of forest disturbances and forest degradation from optical Earth Observation data. Part one comprises an overview of currently available optical remote sensing sensors, which can be used for forest disturbance and degradation mapping. Part two reviews the two main categories of existing approaches: classical image-to-image change detection and time series analysis.

Recent findings: With the launch of the Sentinel-2a satellite and available Landsat imagery, time series analysis has become the most promising but also most demanding category of degradation mapping approaches. Four time series classification methods are distinguished. The methods are explained and their benefits and drawbacks are discussed. A separate chapter presents a number of recent forest degradation mapping studies for two different ecosystems: temperate forests with a geographical focus on Europe and tropical forests with a geographical focus on Africa.

Summary: The review revealed that a wide variety of methods for the detection of forest degradation is already available. Today, the main challenge is to transfer these approaches to high resolution time series data from multiple sensors. Future research should also focus on the classification of disturbance types and the development of robust up-scalable methods to enable near real time disturbance mapping in support of operational reactive measures.

## 1. Introduction

Reliable and operational methods for forest disturbance and forest degradation mapping have become increasingly important for sustainable forest management [1]. A key aspect of sustainable forest management is the monitoring of the forest status, including the assessment of forest disturbances and forest degradation. The term forest disturbance is mostly used for natural causes of crown cover or biomass loss, such as from storm damage, forest fires, drought stress, insect infestations and disease outbreaks but may also include harvesting operations with a potential negative impact. The term forest degradation mostly relates to human-induced crown cover or biomass loss, e.g. for Intergovernmental Panel on Climate Change (IPCC) carbon reporting forest degradation is described as a *"direct human-induced activity that leads to a long-term reduction in forest carbon stocks"* [2]. A further difference between the two terms exists with regard to the temporal impact. A disturbance is usually a single event with a short term impact and may even be regarded as part of the natural forest dynamics, while degradation has a negative long term impact that may be a consequence of one or several single disturbances [3]. Some definitions use forest degradation as an umbrella term for both natural and human-induced forest changes; e.g. the Food and Agriculture Organization of the United Nations (FAO) defines forest degradation as *"changes within the forest which negatively affect the structure or function of the stand or site, and thereby lower the capacity to supply products and/or services"[4].* In this review, we follow the FAO definition and use the term forest degradation for both natural and human-induced changes but we also address methods for mapping disturbances that may not have a long term degrading effect.

Forest degradation mapping by means of remote sensing is essentially a specific application of change detection. Change detection in forest monitoring already has a long tradition starting with Landsat data in the 1980ies and 1990ies [5, 6]. Back then, the main focus was on mapping deforestation and forest regeneration. Assessing and mapping forest degradation is much more challenging than only mapping forest area change. Meanwhile yearly deforestation mapping and the derivation of deforestation rates are already operational at global [7] and also at the national level [8], but there is still only fragmented information available on the extent and magnitude of forest degradation. With Sentinel-2 image data from European Space Agency (ESA) complemented by Landsat 8 from United States Geological Survey (USGS), more high resolution optical data is now available than ever before. These data sets can be used to build a time series. Tracking areas closely over time allows detecting subtle changes (for both deforestation and forest degradation) and generates the possibility to alert in near real time, when changes are occurring.

This paper reviews and discusses available methods used for forest degradation detection from optical remote sensing with a focus on methods based on high resolution data. Methods developed for coarse resolution data e.g. from the 'Moderate-resolution Imaging Spectroradiometer' (MODIS) are only considered if they have the potential to be transferred to high resolution data as well. The reviewed methods can be divided into two categories: i) image to image change detection methods and ii) time series analysis. Due to the increasing availability of high frequency observation data, i.e. more than one image per month, the focus is placed on time-series analysis. Although radar data are another important information source, especially for the tropics, they are not considered for review in this article, as methods and algorithms significantly differ from those applied to optical imagery. The review is complemented by several mapping examples of the authors. Since forest types, seasonal effects as well as degradation drivers differ with geographic location, the mapping examples focus on two main ecosystems: Europe's temperate forests and Africa's tropical evergreen forests.

In temperate forests in Europe, damages caused by storms [9], bark beetles and fires [10] have increased throughout the twentieth century and, are likely to increase further with global warming, though decreasing fires have recently been observed in Mediterranean Europe [11]. Within the first decade of the 21st century, an increase was observed that is substantial enough to be considered a potential threat to the current and future role of European forests as carbon sinks [12]. As a result of this increase, politicians in Europe have recognized the urgent need to gather information on forest health and vitality, and to make this information available to end-users, i.e. forest administration and forest management planning entities. This need is emphasized by the Green paper [13] of the European Commission and in the new European Union (EU) Forest Strategy [1]. Information on degradation caused by storm, wind, snow and human-induced damage by forest operations can be found in the "State of Europe's Forests 2015" report [14], which has been jointly prepared by the signatory countries of "Forest Europe", the former "Ministerial Conference on the Protection of Forests in Europe". However, the usability of this information is limited as it is provided as total area figures per country only and the assessment does not follow a strict nomenclature. While there is a recognized demand for information on forest degradation in Europe, the available data is not yet harmonized and only partly geo-located. Advanced remote sensing technology can provide the timely, accurate and geo-located information that is needed. Such a service is currently being developed by the European Commission (EC) funded DIABOLO project (Distributed, Integrated And Harmonised Forest Information For Bioeconomy Outlooks).

Tropical forests are under even greater pressure than temperate forests: based on a recent review [15], *tropical forests once covered 3.6 billion hectares [...], almost a third have been lost as a result of deforestation.* Of the remaining area, 46% is fragmented, 30% degraded, and only 24% is in a mature and relatively undisturbed state [16]. Some authors indicate even an increase in deforestation in tropical areas in the last years [17, 7]. The main drivers for tropical forest degradation are unsustainable selective logging, forest fires, mining activities and overexploitation of fuel wood [18]. Recent results indicate that tropical forest degradation has a similar or even a greater impact on carbon emissions than tropical deforestation. Depending on the source, emissions from tropical deforestation vary around 8 % [19] (7.4 % [20] and 8.5 % [21]) and from degradation between 6 % [19] and 14% [20]. Forest degradation is therefore a substantial component of overall anthropogenic carbon emissions. The large range of values given for carbon emissions from degradation reflects the difficulty in actually assessing forest degradation [22]. The EC funded project EOMonDis (Bringing Earth Observation Services for Monitoring Dynamic Forest Disturbances to the Users) aims to offer operational Earth Observation (EO) based tropical forest monitoring services to support countries and a wide range of users with accurate relevant forest information data for their management and reporting requirements.

## 2. Optical Earth Observation data sets

The following section presents the characteristics and timelines of different optical high resolution (HR) sensors. In this review, we define 'high resolution' as pixel size between 5 and 30 m. One of the important qualities when choosing the image data sources for a degradation mapping service is the spatial and temporal characteristics of the sensors. While very high spatial (VHR) resolution sensors (< 2 m multispectral pixel size) may have better degradation detection capabilities, the image generally has a much smaller footprint compared to a high or medium resolution sensor. This means that independent of the repeat cycle and revisit frequency of the satellite, the wall to wall coverage using very high resolution sensors is generally more cost and time consuming and in case of larger areas even impossible. Medium to coarse resolution sensors (> 60 m multispectral pixel size), despite having lower cost/time requirements for wall-to-wall data coverages, are very limited in their ability to detect small area disturbances. High resolution satellite systems can be considered a good compromise, offering high enough spatial resolution and large enough footprint for cost-efficient large scale degradation monitoring. Table 1 presents the technical specifications of currently active optical HR satellites that can potentially be used for forest degradation monitoring. With regards to the repeat cycle, the values given for nadir only sensors are ignoring overlaps. The revisit frequency is calculated including possible tilting capabilities of the sensor.

Table 1 Active satellite missions overview - optical HR satellites/sensors specifications

| Satellite system | Mission start/ completion | Spectral characteristics [in µm] | | Orbit height | Swath width | Resolution | Repeat Cycle | Revisit Frequency |
|---|---|---|---|---|---|---|---|---|
| **FormoSat-2 (ROCSat-2)** (NSPO, Taiwan) | 2004 - * | 0.45 – 0.90<br>0.45 – 0.52<br>0.52 – 0.60 | 0.63 – 0.69<br>0.76 – 0.90 | 888 km | 24 km | PAN: 2m<br>Others: 8m | 1 day | 1 day |
| **Landsat 7** (USGS, USA) | 1999 - * | 0.52 – 0.90<br>0.45 – 0.52<br>0.53 – 0.61<br>0.63 – 0.69 | 0.78 – 0.90<br>1.55 – 1.75<br>2.09 – 2.35 | 705 km | 185 km | PAN: 15m<br>Others: 30m | 16 days | 16 days |
| **Landsat 8** (USGS, USA) | 2013 - * | 0.50 – 0.68<br>0.433 – 0.453 0.45 – 0.515<br>0.525 – 0.60<br>0.63 – 0.68 | 0.  845 – 0.885<br>1.36 – 1.39<br>1.56 – 1.66<br>2.10 – 2.30 | 705 km | 185 km | PAN: 15m<br>Others: 30m | 16 days | 16 days |
| **RapidEye** (RapidEye AG, BlackBridge Germany ) | 2008- * | 0.44 – 0.51<br>0.52 – 0.59<br>0.63 – 0.685<br>0.69 – 0.73<br>0.76 – 0.85 | | 630 km | 77 km | 5m | 5.5 days | 1 day |
| **Sentinel-2** (ESA, EU) | 2015 - * (~ 2027) | center wavelength; band width – band :<br>0.443; 0.02 – 1<br>0.490; 0.065 –2<br>0.560; 0.035 – 3<br>0.665; 0.03 – 4<br>0.705; 0.015 – 5 | 0.740; 0.015 – 6<br>0.783; 0.015 – 7<br>0.842; 0.115 – 8<br>0.865; 0.02 – 8a<br>0.945; 0.02 – 9<br>1.375; 0.03 – 10<br>1.610; 0.09 – 11<br>2.190;  0.180 – 12 | 786 km | 290 km | B2,B3,B4,B8: 10m<br><br>B5,B6,B7,B8a, B11,B12: 20m<br><br>B1,B9,B10: 60m | 10 days (Sentinel-2A);<br>5 days (Sentinel-2A & 2B) | 10 days (Sentinel -2A);<br>5 days (Sentinel -2A & 2B) |
| **Spot 6/7** (France) | 2012 - * (~ 2025) | 0.45 – 0.745 0.45 – 0.52<br>0.53 – 0.59<br>0.625 – 0.695<br>0.76 – 0.89 | | | 60 km | PAN: 2.2m<br>Others: 8.8m | 26 days | 1-5 days |
| **UK-DMC-1 / 2** (SSTL, UK) | 2003 - * | 0.52 – 0.62<br>0.63 – 0.69<br>0.76 – 0.90 | | 686 km | 650 km | 32m | 14 days | 1 day |
| **HJ-1A/1B** (China) | 2008 - * | 0.43 – 0.52<br>0.52 – 0.60<br>0.63 – 0.69<br>0.76 – 0.90 | | 650 km | 360 km | 30m | 4 days | 4 days |
| **CBERS-4 (Ziyuan I-04)** (China, Brazil) | 2014 - * | MUXCam:<br>0.45-0.52<br>0.52-0.59<br>0.63-0.69<br>0.77-0.89 | PanMUX:<br>0.51-0.73 (PAN)<br>0.52-0.59<br>0.63-0.69<br>0.77-0.89 | 748 km | MUXCam: 120 km<br>PanMUX : 60 km | MUXCam: 20m<br><br>PanMUX: 5m (PAN), 10m (Others) | 26 days | 3 – 26 days |

| Satellite system | Mission start/ completion | Spectral characteristics [in μm] | | Orbit height | Swath width | Resolution | Repeat Cycle | Revisit Frequency |
|---|---|---|---|---|---|---|---|---|
| **IRS Series, ResourceSat Series** (India) | 1988 - * | PAN: 0.50-0.75 LISS-III: 0.52-0.59 0.62-0.68 0.77-0.86 | 1.55-1.70 LISS-IV: 0.52-0.59 0.62-0.68 0.77-0.86 | Variable | PAN/LISS-IV: 70 km LISS-III: 140 km | PAN/LISS-IV: 5.8m LISS-III: 23.5m | 22-24 days | 5 days |
| **Terra ASTER** (USA, Japan) | 1999 - * | 0.52 - 0.60 0.63 - 0.69 0.76 - 0.86 0.76 - 0.86 | 1.60 - 1.70 2.14 - 2.185 2.185 - 2.225 2.235 - 2.285 2.295 - 2.365 2.360 - 2.430 | 705 km | 60 km | 15m (VNIR), 30m (SWIR) | 16 days | 16 days |

The table shows the list of potential candidate satellite sensors that are collecting optical EO data at high resolution and these sensors enable a new dimension of monitoring capabilities in a multi sensor approach [23], [24], [25]. A regular nadir acquisition scheme is an advantage for a regular and continuous monitoring system. Since the satellites from the Landsat series and Sentinel-2 fulfil this condition and also provide open and cost free access to archives and newly acquired images, these satellites are the workhorses for degradation mapping. The spectral capabilities of the Landsat and Sentinel-2 satellite missions complement each other (Figure 1) and, therefore, it should be possible to increase the density of the time series data by integrating the sensors.

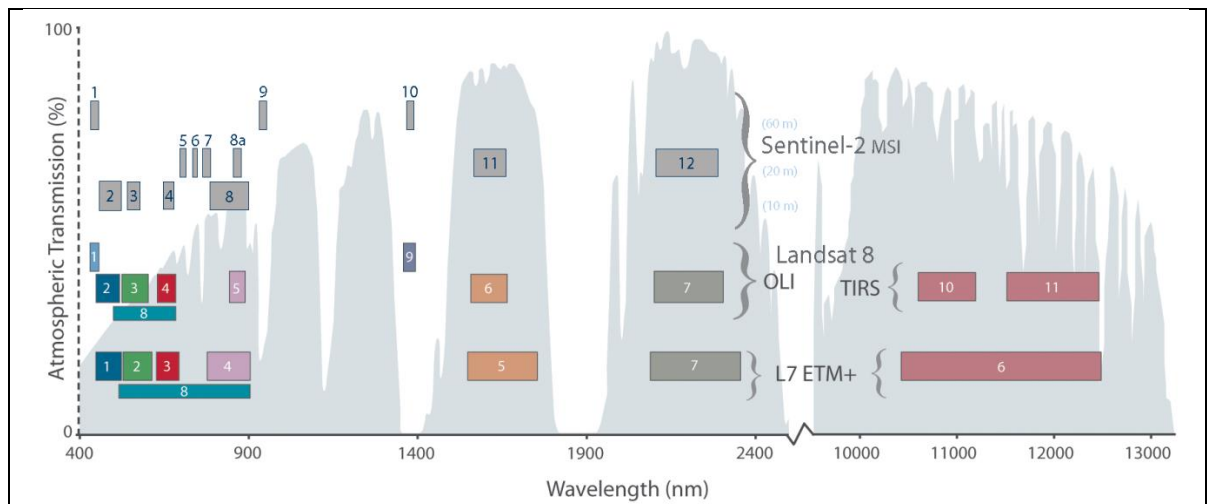

Figure 1 Comparison of Landsat 7 and 8 bands with Sentinel-2 (source: http://landsat.gsfc.nasa.gov)

The Landsat program has been providing continuous multispectral data since 1972 and this data is available without restrictions from USGS. In addition to the Landsat Level 1 standard data products, higher level science data products (e.g. surface reflectance) can also be ordered through a number of data access sites (URLs: http://earthexplorer.usgs.gov, last accessed 29.12.2015; http://glovis.usgs.gov, last accessed 29.12.2015). The current operational satellites are Landsat 7 and Landsat 8. It should be mentioned, that Landsat 7 ETM+ has had a failure of the Scan Line Corrector (SLC) leading to missing data in all images from 2003 onwards and therefore some limitations in usability of this data. The long term continuity of data from the program is foreseen to continue well into the future, with Landsat 9 planned for launch in 2023, although there is a risk, that Landsat 8 could stop working before Landsat 9 is in orbit.

Sentinel-2 is planned as a "two satellite mission": the first, Sentinel-2A, was launched June 23[rd], 2015 (full operational readiness is planned for July 2016) and Sentinel-2B is planned for launch in 2017. The operational lifespan of the Sentinel-2 mission is 7.25 years, while the consumables can last for up to 12 years (source: ESA, https://earth.esa.int/web/guest/missions/esa-operational-eo-missions/sentinel-2, accessed 22.12.2015). In order to guarantee continuity with two parallel sensors in orbit, ESA has already contracted Airbus Defense and Space for the construction of Sentinel-2C and -2D. Available Sentinel-2 images can be downloaded at the Scientific Data Hub and have been provided in the form of a "rolling archive" since December 2015.

## 3. Methods for forest degradation mapping

The methods for forest degradation mapping from optical HR image data can be subdivided in two ways: First, in terms of type and number of data sets used; and second, in terms of features to be mapped. These features can be mapped directly from the digital numbers or surface reflectance values from the image bands or indirectly through indices. Commonly used indices are for example: normalized difference vegetation index (NDVI), normalized burnt ratio (NBR), enhanced vegetation index (EVI) and the soil-adjusted vegetation index (SAVI). In the following matrix the references cited in this review are subdivided according to the methods and features they use for degradation mapping. Sometimes, methods are a mixture or a combination of different methods. In this case, they are mentioned in both categories. Also the mapping examples (MEs) given in Section 4 are listed in the table.

Table 2: Reviewed papers divided by methods and features used

| Type & number of images → | Image to image change detection | Time series analysis based change detection | | | |
|---|---|---|---|---|---|
| Time series subcategories → ↓ Features (direct or indirect) | (bi- or multitemporal) | a) Threshold-based classification | b) Curve fitting | c) Trajectory fitting | d) Trajectory segmentation |
| Crown cover change | [26], [27], [28] | [29], [30] [31], | [32], [33], ME2 [34] | [35] | [36, 37], [38] |
| Defoliation, leaf discoloration | ME1 [39] | [40], [31], ME1 [39] | | [35], [41] | [40], [42] |
| Logging roads, skid trails, gaps | [43], [28], [44], [45] [46] | [29], [46], [68] | [47] | ME4, ME5 [48] | [36], [42], [38] |

One simple method for degradation mapping is to classify only one image and compare it against an existing map or against the assumption of complete intact forest. This method is not change detection in a strict sense, as only one remote sensing image is involved, ME3 illustrates this approach. The main two categories of change detection approaches are:

- Classical image to image change detection - At least one image acquired before and one after a degradation event is required, of which the first must be from the start of a monitoring period and the second at the end. This approach is often also referred to as "bi temporal change detection" or just "change detection". Such an approach can be also applied two or more times, each time comparing two images or classifications.

- Time series analysis based change detection – Requires a series of images taken continuously over a period of time. Thus, there is a need for substantially more and regular image acquisitions over the area of interest. This approach is often referred to simply as "time series analysis".

## 3.1. Image-to-image change detection

Image to image change detection is currently the most commonly used method for mapping forest degradation. It makes use of the change of the spectral signatures of the land surface between the images taken at two different dates in time in order to assess changes that occurred between these two dates. For image to image change detection a forest mask is necessary in order to focus on changes in those forest areas only, since other major land cover categories, especially agricultural areas, can change significantly over the course of one vegetation period and would lead to false results. Such forest masks must be as up-to-date as possible and can be either derived from the EO data used for change detection or from other sources. Amongst the standard pre-processing steps, geometric and radiometric calibration is highly important. Compared to absolute calibration, relative calibration provides the advantage that seasonal changes can be better separated from real changes when the images utilized are from different stages of the seasonal development. Changes are then classified based on the change of indicators between the two dates, using indicators such as the NDVI or the NBR. Numerous approaches to identify an optimal indicator for the detection of degradation have been developed and tested. These include approaches that use reflectance values from the available image channels of both images to derive a change metric. These indicators can be used in a threshold approach or a standard classification procedure, as well as in combination with classification approaches that integrate image segmentation. A stack of different image channels from the two images and/or change indicator images can also be used for visual enhancement of the classification based results. There are, thus, a variety of methods that have developed to a level of maturity that allow operational applications.

Comprehensive overviews of the change detection mapping methods have been published over the last decade [49], [50], [51], [52] and [12]. Some state-of-the-art change detection methods are presented and discussed in a recent review paper based on publications from 2007 to 2013 [53]. Negative crown cover changes can be seen as a feature indicating forest degradation. Therefore, crown cover change maps are frequently derived for degradation monitoring in tropical areas, e.g. for India [27]. Crown cover changes were also an important feature for detecting annual forest degradation in Mato Grosso, Brazil, between 1992 and 2004 [28]. An approach based on a supervised classification of Landsat data determines the tree cover density for defined epochs and then calculates tree cover changes between individual epochs [26]. Mapping gaps, logging roads, skid trails and burnt areas as features for degradation has also been used frequently for tropical areas [44, 45, 28]. In Europe, some of these feature detection methods have also been used for detecting forest fire damages, storm damages and biotic damages, e.g. in the frame of the EUFODOS project [43].

## 3.2. Time series based change detection

A typical remote sensing time series consists of three components: a long term directional trend component, a seasonal component and a residual component, see Figure 2 [54]. Depending on the research focus, either one or all of these components, are of special interest [41]. A widely used tool to extract the individual components is the BFAST tool (Breaks For Additive Season and Trend, [41]). Similarly, there are several options for separating components and smoothing the time series available in TIMESAT [55] with a wide variety of applications. TIMESAT integrates a number of processing steps to transform the noisy signals into smooth seasonal curves including e.g. the Savitzky-Golay filter.

With regards to forest degradation monitoring, the residual component is of high interest and has to be differentiated from residual noise. Depending on the forest ecosystem region, also the seasonal component plays an important role in the characterization of the forest disturbances.

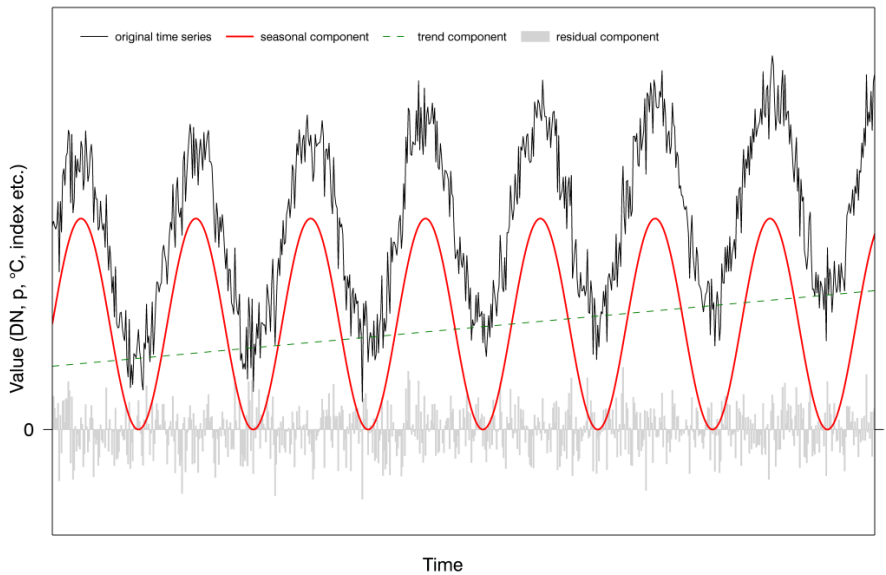

Figure 2: Components of time series (from [54])

Time series data, approaches, opportunities and challenges are reviewed in a chapter [23] of a whole book on time series applications from both optical and radar data sources [54]. Time series approaches can be based on series of raw digital numbers (DN), reflectance values (%) or different variables. These variables can be geophysical variables, e.g. top of atmosphere (TOA), Photosynthetic Active Radiation (PAR), Fraction of absorbed Photosynthetic Active Radiation (FPAR), index variables such as NDVI, Enhanced Vegetation Index (EVI), Soil Adjusted Vegetation Index (SAVI), thematic variables (classes or spectral unmixing results), topographic variables (e.g. slope, aspect, height) or texture variables (e.g. homogeneity, object size and shape, connectivity) [54]. A pre-requisite for application of time series methods is that the image data is precisely pre-processed, both geometrically as well as radiometrically. Geometric pre-processing is usually performed by fully automated image co-registration using sub-pixel image matching algorithms [56]. For radiometric pre-processing of time series data, there are two options: absolute and relative calibration. A comparison between absolute and relative radiometric correction was done for Landsat data with similar results [57]. For absolute atmospheric calibration, the idea is to move from digital numbers to physical surface reflectance values (bottom of atmosphere values). Parameters for the atmospheric conditions in the different parts of the image are needed to perform this calibration. In contrast, relative radiometric calibration does not need additional input data, as the images are relatively adjusted to each other or a master scene. For classical change detection, relative radiometric calibration was widely used. For time series, the standard procedure is to apply an absolute atmospheric correction [58] in order to add new data sets to the time series immediately upon availability. Tools for atmospheric correction of optical data are continuously being improved by organizations such as ESA (Sen2Cor for Sentinel2 imagery [59]) or USGS (LEDAPS). LEDAPS means Landsat Ecosystem Disturbance Adaptive Processing System and is freely available at the USGS website (URL: http://landsat.usgs.gov). Landsat 8 data can also be downloaded in reflectance values without any further processing. In addition, there are also other (relative radiometric adjustment) tools available such as the 'Multitemporal and Multispectral Method to Estimate Aerosol Optical Thickness' [60]. Further advances in preprocessing focus on topographic correction [61] and on cloud mask generation [62] [63], [64].

Various time series analysis methods have been developed over the last few decades for long-term vegetation monitoring based on low to medium resolution satellite imagery such as AVHRR and MODIS. An operational system used for forest degradation mapping is the ForWarn system [31], which provides near real time forest monitoring based on MODIS time series. A similar system for deforestation monitoring (DETER) is operationally in place in Brazil [65, 66]. An example for forest fire detection and monitoring of the fire affected areas based on a MODIS time series analysis on the Peloponnese in Greece is shown in Figure 3.

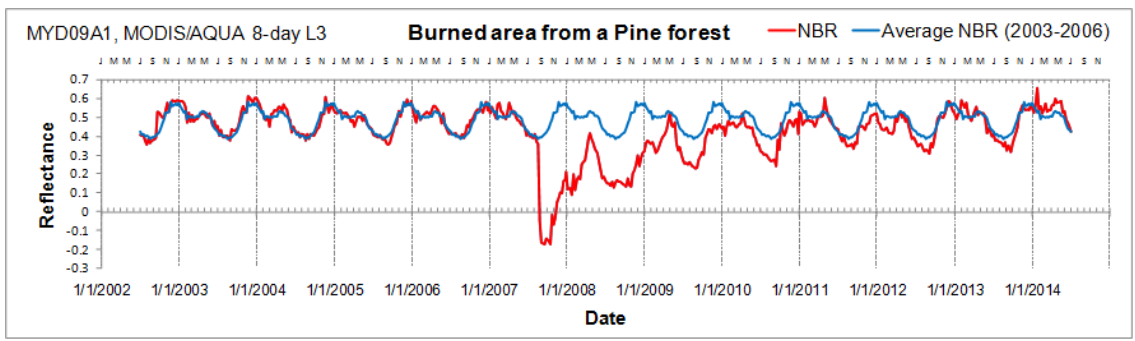

Figure 3: Time series from MODIS data showing a forest fire and the vegetation recovery (Greece)

All applications on medium resolution data have one important limitation: whereas the temporal resolution of MODIS data is high, the spatial resolution of 250m per pixel only allows monitoring severe large-scale degradation, such as from e.g. large forest fires or massive biotic damages. The typical size of degradation patches however is often much smaller than the MODIS pixel resolution and therefore degradation patches cannot be detected. This limitation was also pointed out in a review paper for tropical forest monitoring in Asia [67]. Despite the limitations in terms of resulting maps, methods may still be transferable to HR data as well and are therefore included in this review.

Time series data for Sentinel-2 or Landsat-8 typically show much lower densities than e.g. for MODIS data and the availability of time series data from HR satellites strongly depends on regional cloud cover and geographic latitude. In the coming years, the Landsat and Sentinel-2 missions will provide the first free and dense HR time series data covering multiple years. Advances in image processing and an increase in computational capacity have stimulated the development and application of time series methods for forest monitoring based on high resolution time series. A review on forest monitoring methods based on Landsat time-series [53] concluded, that many software related issues and challenges still remain. These challenges need to be solved in order to harness the full capabilities offered by rich, freely available Landsat and Sentinel-2 time series datasets. In particular, software related issues associated with pre-processing, analysis and validation require further research.

Time series analysis methods can be divided into four sub-categories: (a) threshold based change detection; (b) curve fitting; (c) trajectory fitting and (d) trajectory segmentation [53]. A schematic comparison of the four methods is given in Figure 4.

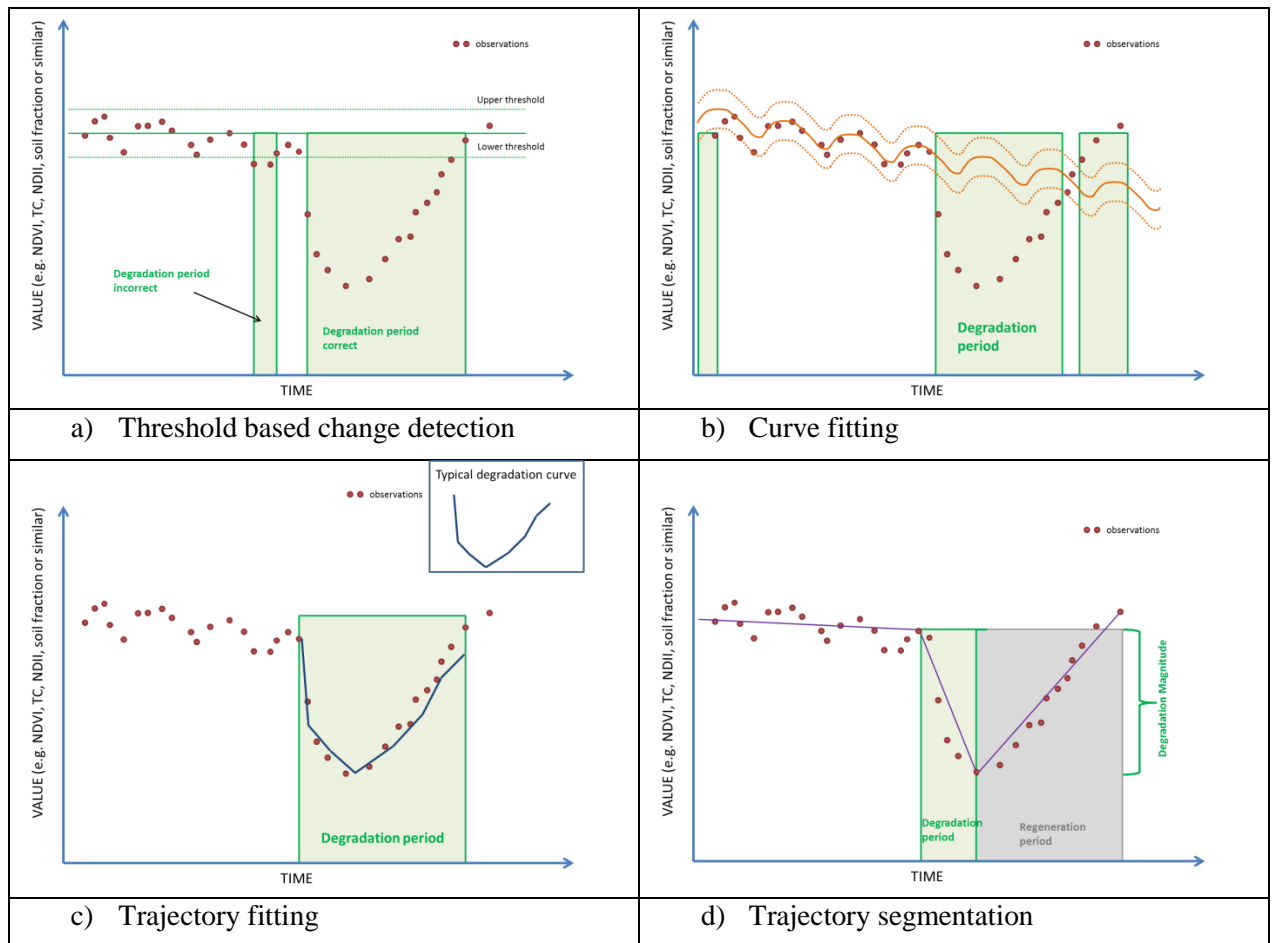

Figure 4: Schematic illustration of different time series analysis methods

a)  Threshold based change detection

These methods use a thresholding procedure to separate forest from non-forest or intact from degraded forest in a time series. The already mentioned ForWarn System [31] basically belongs to this group of approaches. An automated forest change mapping algorithm (VCT) was e.g. developed based on the likelihood of a pixel being forested [29]. For forest and non-forest, a threshold of a so called Integrated Forest Z-score is applied to identify forested areas. Another comparable threshold based approach [30] was used to map changes in forested wetland ecosystems based on tasselled cap images. A further thresholding approach was implemented for mapping forest fires in the Mediterranean area [39]. The main drawback of threshold-based time series analysis is the same as with any threshold based techniques: the thresholds are empirically determined and thus cannot be directly transferred to other study areas, which are characterized by different vegetation types, vegetation densities or degradation patterns [53]. Based on MOD13 data sets, insect defoliation in Norwegian pine forests were mapped by combining both threshold-based methods with some trajectory segmentation attempts in order to overcome the limitations of procedures solely based on thresholding [40]. A specific method combining both the image to image approach with time series has recently been used in support of Global Forest Watch [46]. They use a classification tree algorithm to compare new images to metrics of a time series in a very straightforward approach for tropical evergreen forest with an example mapping result for the Republic of Congo. In Eastern Europe, a similar approach was followed [68] based on previous work [69].

### b) Curve fitting

Several studies demonstrated the utility of curve fitting approaches for monitoring forest dynamics. Often, seasonal effects are suppressed by selecting near-anniversary date images (e.g. acquisitions at peak vegetation growth). A pixel-wise trend function based on regression models is fitted between spectral variables (SV), which can be single band reflectance or derived indices, and time [32]. A slope that is significantly different from zero determines presence or absence of a trend. The trend is expressed as the slope of the regression curve and provides information on the magnitude of change over time. The sign of the slope can be used to separate different change processes such as increase of crown cover or decrease of crown cover. Linear and quadratic coefficients from Landsat time series were used to analyze forest cover change trends [33]. In addition to the slope, as a measure to summarize the linear trend, the coefficients of quadratic curvature were found to be important for understanding degradation and recovery. Such an approach was also followed for mapping tropical forest degradation in Brazil [47], where the relation of forest phenology and selective logging based on a time series using MODIS data has been investigated. Aside from using indices, such as enhanced vegetation index (EVI), also a spectral mixture analysis was applied. A known shortcoming of single curve fitting, as described above, is that standard underlying statistical assumptions, such as data normality and equal variance, are generally difficult to meet [70] and violation of these assumptions results in an inadequate representation of the data by the fitted function. Functional time series analysis as a statistical approach for curve fitting has been applied for monitoring storm damages and forest fires based on dense MODIS time series [34]. In addition to deriving the trend component, curve fitting also captures the underlying dynamic of seasonality.

### c) Trajectory fitting

Many degradation patterns in forests have a distinct temporal behavior before and after the degradation event, resulting in a "*characteristic spectro-temporal signature*" [35]. Whereas a forest experiencing gradual change might be represented by a linear trend, a stable forest (i.e. no degradation) can be represented by a horizontal line. Similarly, abrupt changes may result in an abrupt change in the trajectory followed by stable, increasing or decreasing trends, depending on the specific forest degradation and recovery processes [53]. Thus, the changes are analyzed by fitting idealized trajectories, which are characteristic for specific degradation types. This requires the definition of hypothesized change trajectories which are based on the expected spectral-temporal behavior of selected variables. These hypothesized trajectories have to be defined for each degradation type separately. Trajectory fitting can therefore be interpreted as "*a supervised change detection method with idealized trajectories representing training signatures specific to different degradation types*" [41, 53]. This approach was e.g. applied for monitoring degradation in tropical forests [48]. The main challenges for this application were the atmospheric conditions in the Tropics, the fast regrowth of vegetation after selective logging, and the small-structured logging patterns. The major limitation of the trajectory fitting approach is that the shape of the typical degradation curve(s) must be known from some reference data; and the method will only work properly if the observed spectral trajectory matches one of the predefined typical degradation curves [53].

d) Trajectory segmentation

The so-called LandTrendr approach decomposes the trajectory into a series of straight-line segments to capture broad features of the trajectory as well as sub-trends [37]. The first phase of segmentation is the determination of the vertex years that define the end points of segments. In the second phase the best straight-line trajectory is fitted through those vertices using either point-to-point or regression lines. The result of this segmentation is a spectral trajectory which is composed of straight-line segments. The time position and spectral value of vertices of the segments provide the essential information, which is needed to produce maps of forest change, or serve as predictor variables to compare future developments with. According to [53], *"the advantage of this approach is that the straight-line segments allow the detection of abrupt events, such as disturbances, as well as longer-duration processes, such as regrowth"*. Another advantage is that no typical curve of degradation is required as in the previous method, because the data themselves determine the shape of the trajectory. The LandTrendr algorithm captures abrupt disturbances such as clear-cuts just as well, or even better, than two-date change detection methods and detects subtle changes such as insect-related degradation and growth with reasonable robustness [38]. A main drawback of the method is that seasonal effects caused by phenology are not taken into account. LandTrendR has been used to describe the status of the forests in Europe applying a tasseled cap wetness indicator in an annual time series [36]. The same approach was used for mapping forest disturbances in the U.S. [42].

## 4. Some Forest Degradation Mapping Examples

In the following, five mapping examples are given to illustrate the methods described above. They cover various methods explained above. In order to illustrate the examples, Figure 5 shows some resulting maps from these mapping examples.

### 4.1. European forest monitoring examples

ME1: Threshold based change detection for mapping forest fires
The mapping example presented here uses a semi-automatic method to map burned surfaces by comparing multi-temporal Landsat images using a set of spectral-based rules [39]. If a rule fails to correctly capture the burned area then it is possible either to adjust the thresholding coefficients of the rule or to discard the rule. This method, being free from the training phase of the algorithm, minimizes human intervention and therefore, increases the objectivity. This allows the mapping algorithm to run for a series of satellite images that would otherwise need a lot of processing time. Both these issues are important when many satellite images need to be processed, such as for the spatially explicit reconstruction of recent fire history where hundreds of images might be used in the processing chain.

When the method was applied to two case studies in Mt. Parnitha and Samos Island in Greece then the overall accuracy was 95.69% and 93.98%, respectively, while the commission and omission errors were 6.92% and 10.24% for Mt. Parnitha and 3.97% and 8.80% for Samos Island, respectively. The semi-automatic rule-based method was applied to a series of USGS Landsat images from six different scenes that covered Attica and the Peloponnese in Greece of the periods 1984-1991 and 1999-2009. In total 1773 fires were identified and mapped with the majority of non-mapped fire scars corresponding to size classes of 0-1 ha and 1-5 ha, where the loss in fire scar mapping is high. However, this is expected since small fires can be identified and recorded by forest authorities, but cannot be captured by satellite remote sensing technology due to spatial, spectral and temporal restrictions imposed by the satellite sensor used.

ME2: Curve fitting for mapping forest damages in Europe from MODIS dense time series

In this mapping example, three different statistical methods for curve fitting are compared: spatial; temporal and combined spatio-temporal forecasts. Spatial forecast means, that a clustering algorithm is used to group the time series data based on the features normalized difference vegetation index (NDVI) and the short wave infrared band (SWIR). For estimation of the typical temporal behavior of the NDVI and SWIR during the vegetation period of each spatial cluster, several methods of functional data analysis including functional principal component analysis were applied. The temporal forecast (i.e. the curve) is carried out by means of functional time series analysis and an autoregressive integrated moving average model. The combination of the temporal forecasts and spatial forecasts lead to the spatio-temporal forecast. The main pre-requisite for applying such approaches is a really dense time series, which was so far only available from coarse resolution data. Therefore this mapping example is utilizing MODIS data. Two study areas were selected for testing and evaluation of the methods: Study area 1 is located in the south of Germany where storm damages occurred in the monitoring period. Study area 2 is located in northern Spain where forest fires frequently occur. For the spatio-temporal approach, overall accuracies of 76% in the Spanish and 94% in the German study site were achieved [34].

## 4.2. Tropical forest monitoring examples in Africa

ME3: Classification vs assumed intact forest for mapping degradation in the Democratic Republic of Congo (DRC)

In the ReCover project (FP7 GA No 263075) in a demonstration case for Democratic Republic of Congo, a static approach using non-parametric support vector machine (SVM) based classification of RapidEye data was used to map degradation as reduced forest canopy cover compared to fully canopy covered intact forest. While many studies focus on identifying forest degradation as a change over time, the objective of this study was to test the performance of SVM classifier as a tool to map forest degradation using only one optical dataset. The RapidEye image at a ground resolution of 5m was used as reference to create a training dataset where forest, non-forest and degraded forest polygons were used to generate random training points (> 30000 samples). Classification was performed using SVM algorithm in open source R-statistical package [71] and a 100 fold bootstrapping was used to randomly draw samples to build the classification model [72]. From the classified image, the degraded forest areas were extracted using the assumption that a hectare of forest area is degraded if > 20% of the pixels correspond to degraded forest area. The bootstrapping was also used for accuracy assessment and led to an overall accuracy of 0.91 for the three classes forest, non-forest and degraded forest. Producer's and user's accuracy for the class degraded forest is 0.83 and 0.92 respectively. The values are expected to be lower, if an independent accuracy assessment would be performed.

ME4: Trajectory fitting for degradation mapping in the Republic of Congo

The aim of this application was to map REDD+ activity data, i.e. degradation assessment by separating degraded from intact forests. The basis for the assessment was a dense time series of satellite imagery consisting of Spot-HR, Landsat 5-8, RapidEye, Pléiades, ALOS AVNIR and ASTER data. For geometric and radiometric pre-processing, RSG and IMPACT software tools developed by Joanneum Research were used which allow efficient processing of a large number of satellite scenes. After the pre-processing steps, spectral mixture analysis is performed to estimate the fractions of "shadow", "green vegetation" and "soil". Whereas the soil fraction signal diminishes rapidly after degradation (e.g. within several months after degradation, because of rapid re-vegetation), the signal from the green vegetation fraction shows long term changes over several years. To utilize the rich information content and to reduce noise inherent in the dense time series, a trajectory fitting approach was applied. As basis, typical temporal development curves of the "shadow", "green vegetation" and "soil" fractions were derived for disturbed / degraded and intact forest areas. These typical temporal trajectories are moved along the time-axis to each acquisition date. For each acquisition date, then the differences between the observed fractions (from the spectral mixture analysis) and the end-member fractions according to the typical trajectories are calculated. In the next step, each pixel is assigned to the trajectory which best fits to the observations. The pixel-based results were then generalization to a minimum mapping unit of 1 ha, which significantly reduces noise in the results. As the last processing step, a visual refinement was performed e.g. to exclude false change indications which frequently occurred in swamp-forests (caused by varying wetness-conditions). For quality control, a plausibility check with stratified sampling was performed. The forest degradation map was used for stratification, where random sampling points were selected within the intact forest stratum (1056 points) and within the mapped degradation areas (94 points). On-screen visual interpretation was then performed based on the time-series imagery, complemented with higher resolution imagery from Rapideye, IKONOS, Worldview and Pleiades for parts of the test area. As forest degradation occurs at very low rates (e.g. detected in 1.5% of the forests over a ten-year period only), the user and producer accuracies of the intact forest area is naturally very high. The plausibility check shows producer accuracies of 99.6% for intact forest and 90% for degraded forest and user accuracies of 99.9% for intact forest and 75.1% for degraded forest. These figures might overestimate the actually achieved accuracies, as they are not derived from a fully independent accuracy assessment. The work was performed within the project GSE-Forest Monitoring REDD Extension Services, financed by ESA.

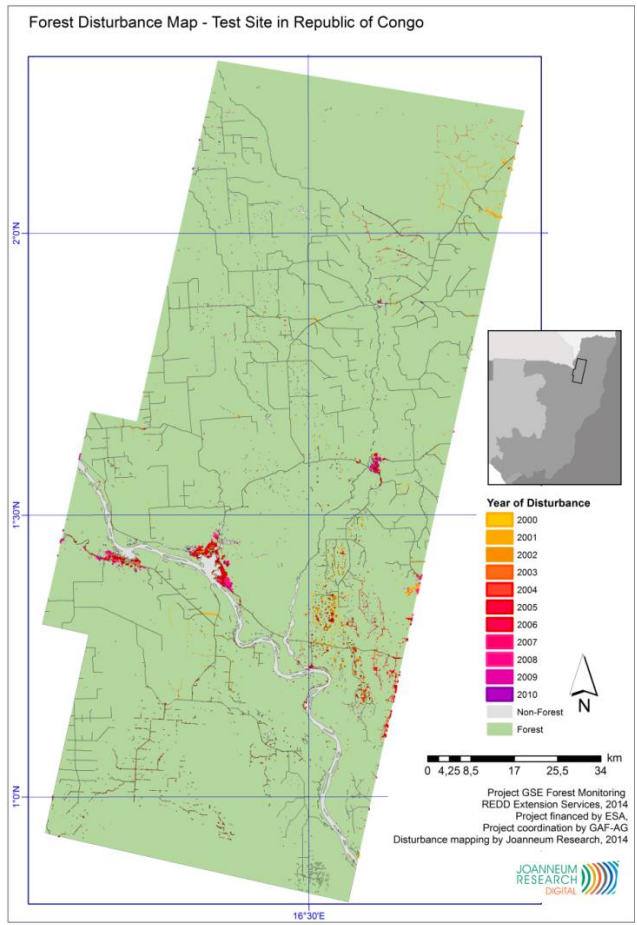

Figure 5: Application example for mapping forest degradation within the Republic of Congo.

ME5: Trajectory fitting for degradation mapping in Cameroon and the Central African Republic

In the frame of REDD+, forest monitoring by Earth Observation plays an important role. Within the EC-funded research project REDDAf (FP7 GA No 262775), forest degradation has been mapped in two test sites in Cameroon and in the Central African Republic [48]. The main driver for forest degradation in these areas is selective logging. Remote sensing based detection of forest degradation from selective logging is difficult, as these subtle degradation signals are not easy to detect in the first place and quickly lost over time due to fast regrowth. To overcome these shortcomings, a time series analysis using a trajectory fitting method has been used to map forest degradation over roughly a decade with minimum annual image updates of Landsat data. The approach involves a series of pre-processing steps like geometric and radiometric adjustments followed by feature selection procedure. Twelve features were tested for their usability: individual bands, various indices (NDVI, TVI, SAVI, NDII5/7) and further soil fraction and mNDFI from spectral mixture analysis. Soil fraction showed the best results compared to visual interpretation of gaps and was therefore used for the classification of spectral curves. The resulting pixel-based classification, representing the observed individual logging gaps, was then aggregated to potential degradation zones. The method was developed on a test site in Cameroon and applied to an area in the Central African Republic. For both areas, the results were finally evaluated against degradation areas derived from visual interpretation of VHR optical imagery. Results showed overall accuracies in both test sites above 85%.

## 5.    Conclusions

The review of methods shows that there are already many methods available for bi-temporal change detection from high resolution data on the one hand and for time series analysis from coarse resolution data on the other hand. The current main challenge and research development focus is transferring these approaches to high resolution time series data which is currently becoming available from Sentinel 2 (in combination with Landsat data) and to improve the preprocessing quality of the high resolution time series. Further these methods need to be developed towards disturbance classification to enable the analysis of the different disturbance types such as e.g. fire, storm, insect caused damages, degradation by selective logging etc. along with the increasing density and length of time series. This has to be accompanied by a focus on the development of robust up- scalable methods that will enable both near real time disturbance mapping in support of operational reactive measures and as well the development of long term regional and global observation capacities for disturbances by major disturbance types.

## 6.    Acknowledgements

This project has received funding from the European Union's Horizon 2020 research and innovation programme under grant agreement No 685761 (EOMonDIS) as well as under grant agreement No 633464 (DIABOLO). The study in Congo was performed within the project GSE-Forest Monitoring REDD Extension Services, financed by ESA.

**References:**

[1]     European Commission, "Communication from the Commission to the European Parliament, the Council, the European Economic and Social Committee and the Committee of the Regions: A new EU Forest Strategy: for forests and the forest-based sectors." online, last access: 11 April 2016 2013.

[2]     IPCC, "Good Practice Guidance for Land Use, Land-Use Change and Forestry (GPG-LULUCF)." http://www.ipcc-nggip.iges.or.jp/public/gpglulucf/gpglulucf_contents.html, 2003. accessed 5 Oct 2016.

[3]     FAO, "Forest Resources Assessment Working Paper 177: Assessing forest degradation: Towards the development of globally applicable guidelines." http://www.fao.org/docrep/015/i2479e/i2479e00.pdf, Rome, 2011.

[4]     D. Schoene, W. Killmann, H. Lüpke, and M. LoycheWilkie, "Forest and Climate Change Working Paper 5: Definitional issues related to reducing emissions from deforestation in developing countries." ftp://ftp.fao.org/docrep/fao/009/j9345e/j9345e00.pdf, 2007. last accessed: May 25th, 2016.

[5]     A. Singh, "Review article digital change detection techniques using remotely-sensed data," *International Journal of Remote Sensing*, vol. 10, pp. 989–1003, jun 1989.

[6]     P. R. Coppin and M. E. Bauer, "Change Detection in Forest Ecosystems with Remote Sensing Digital Imagery," *Remote Sensing Reviews*, no. 13, pp. 207–234, 1996.

[7]     M. C. Hansen, P. V. Potapov, R. Moore, M. Hancher, S. A. Turubanova, A. Tyukavina, D. Thau, S. V. Stehman, S. J. Goetz, T. R. Loveland, A. Kommareddy, A. Egorov, L. Chini, C. O. Justice, and J. R. G. Townshend, "High-resolution global maps of 21st-century forest cover change," *Science*, vol. 342, no. 6160, pp. 850–853, 2013.

[8]     P. Potapov, S. Turubanova, M. Hansen, B. Adusei, M. Broich, A. Altstatt, L. Mane, and C. Justice, "Quantifying Forest Cover Loss in Democratic Republic of Congo, 2000-2010, with Landsat ETM+ data," *Remote Sensing of Envionment*, vol. 122, pp. 106–116, 2012.

[9]     B. Gardiner, K. Blennow, J.-M. Carnus, M. Fleischer, F. Ingemarson, G. Landmann, M. Lindner, M. Marzano, B. Nicoll, C. Orazio, J.-L. Peyron, M.-P. Reviron, M.-J. Schelhaas, A. Schuck, M. Spielmann, and T. Usbeck, "Destructive storms in European forests: past and forthcoming impacts." http://ec.europa.eu/environment/forests/pdf/STORMS%20Final_Report.pdf, 2010. Last accessed: Oct, 5th 2016.

[10]    N. Koutsias, G. Xanthopoulos, D. Founda, F. Xystrakis, F. Nioti, M. Pleniou, G. Mallinis, and M. Arianoutsou, "On the relationships between forest fires and weather conditions in Greece from long-term national observations (1894-2010)," *International Journal of Wildland Fire*, vol. 22, no. 4, pp. 493–507, 2013.

[11]    M. Turco, J. Bedia, F. Di Liberto, P. Fiorucci, J. von Hardenberg, N. Koutsias, M.-C. Llasat, F. Xystrakis, and A. Provenzale, "Decreasing Fires in Mediterranean Europe," *PLoS ONE*, vol. 11, p. e0150663, March 2016.

[12]    • R. Seidl, M.-J. Schelhaas, W. Rammer, and P. J. Verkerk, "Increasing forest disturbances in Europe and their impact on carbon storage," *Nature Climate Change*, vol. 4, pp. 806–810, August 2014. **On the basis of an ensemble of climate change scenarios, the authors find that damage from wind, bark beetles and forest fires in Europe is likely to increase further in coming decades, and they estimate the rate of increase to be +0.91 × 10⁶ m³ of timber per year until 2030.**

[13]    European Commission, "Green Paper: On Forest Protection and Information in the EU: Preparing forests              for              climate              change."              http://eur-lex.europa.eu/LexUriServ/LexUriServ.do?uri=COM:2010:0066:FIN:EN:PDF, 2010. Last accessed: Oct, 5th 2016.

[14]    "Full State of Europe's Forests 2015." http://www.foresteurope.org/fullsoef2015, 2015. Last accessed: Oct. 5ht 2016.

[15]    International Sustainability Unit, "Tropical Forests - A Review." http://www.pcfisu.org/wp-content/uploads/2015/04/Princes-Charities-International-Sustainability-Unit-Tropical-Forests-A-Review.pdf, 2015. Last accessed: Oct, 5th 2016.

[16]    L. L. Susan Minnemeyer, N. Sizer, C. Saint-Laurent, and P. Potapov, "A world of opportunity." http://www.wri.org/sites/default/files/world_of_opportunity_brochure_2011-09.pdf, 2011. Last accessed: Oct, 5th 2016.

[17]    D.-H. Kim, J. O. Sexton, and J. R. Townshend, "Accelerated deforestation in the humid tropics from the 1990s to the 2000s," *Geophysical Research Letters*, vol. 42, pp. 3495–3501, may 2015.

[18]    R. DeFries, F. Achard, S. Brown, M. Herold, D. Murdiyarso, B. Schlamadinger, and C. de Souza, "Earth observations for estimating greenhouse gas emissions from deforestation in developing countries," *Environmental Science & Policy*, vol. 10, no. 4, pp. 385–394, 2007.

[19]    N. L. Harris, S. Brown, S. C. Hagen, S. S. Saatchi, S. Petrova, W. Salas, M. C. Hansen, P. V. Potapov, and A. Lotsch, "Baseline map of carbon emissions from deforestation in tropical regions," *Science*, vol. 336, pp. 1573–1576, June 2012.

[20]    R. A. Houghton, J. I. House, J. Pongratz, G. R. van der Werf, R. S. DeFries, M. C. Hansen, C. L. Quéré, and N. Ramankutty, "Carbon emissions from land use and land-cover change," *Biogeosciences*, vol. 9, pp. 5125–5142, dec 2012.

[21]    J. Grace, E. Mitchard, and E. Gloor, "Perturbations in the carbon budget of the tropics," *Global Change Biology*, vol. 20, pp. 3238–3255, June 2014.

[22]    J.-P. Lanly, "Deforestation and forest degradation factors," in *Proceedings of XII World Forestry Congress*, 2003.

[23]    • C. Kuenzer, S. Dech, and W. Wagner, eds., *Remote Sensing Time Series: Revealing Land Surface Dynamics*. Springer International Publishing, 2015. **Comprehensive review of different time series approaches for different data sets and applications from around the globe.**

[24]    M. Wulder, T. Hilker, J. White, N. Coops, J. Masek, D. Pflugmacher, and Y. Crevier, "Virtual constellations for global terrestrial monitoring," *Remote Sensing of Environment*, vol. 170, pp. 62–76, 2015.

[25]    P. Hostert, P. Griffiths, S. van der Linden, and D. Pflugmacher, *Time Series Analyses in a New Era of Optical Satellite Data*, ch. 15: Forest Cover Dynamics During Massive Ownership Changes - Annual Disturbance Mapping Using Annual Landsat Time-Series, pp. 307–322. Springer International Publishing, 2015.

[26]    B. A. Margono, S. Turubanova, I. Zhuravleva, P. Potapov, A. Tyukavina, A. Baccini, S. Goetz, and M. C. Hansen, "Mapping and monitoring deforestation and forest degradation in Sumatra (Indonesia) using Landsat time series data sets from 1990 to 2010," *Environmental Research Letters*, vol. 7, pp. 1–16, 2012.

[27]    N. H. Ravindranath, N. Srivastava, I. K. Murthy, S. Malaviya, M. Munsi, and N. Sharma, "Deforestation and Forest Degradation in India - Implications for REDD+," *Current Science*, vol. 102, pp. 1117–1125, April 2012.

[28]    E. A. Matricardi, D. L. Skole, M. A. Pedlowski, W. Chomentowski, and L. C. Fernandes, "Assessment of tropical forest degradation by selective logging and fire using Landsat imagery," *Remote Sensing of Environment*, vol. 114, pp. 1117–1129, 2010.

[29]    C. Huang, S. Goward, J. Masek, N. Thomas, Z. Zhu, and J. Vogelmann, "An automated approach for reconstructing recent forest disturbance history using dense Landsat time series stacks," *Remote Sensing of Envionment*, vol. 114, pp. 183–198, 2010.

[30]    N. Kayastha, V. Thomas, J. Galbraith, and A. Banskota, "Monitoring wetland change using inter-annual Landsat time-series data," *Wetlands*, vol. 32, pp. 1149–1162, 2012.

[31]    W. Hargrove, J. Spruce, G. Gasser, L. Martin, and S. Norman, "Monitoring Regional Forest Disturbances across the US with Near Real Time MODIS NDVI Products included in the ForWarn Forest Threat Early Warning System." Presented at the 2013 AGU Fall Meeting, 2013.

[32]    J. E. Vogelmann, G. Xian, C. Homer, and B. Tolk, "Monitoring gradual ecosystem change using Landsat time series analyses: Case studies in selected forest and rangeland ecosystems," *Remote Sensing of Environment*, vol. 122, pp. 92–105, 2012.

[33]    E. Lehmann, J. Wallace, P. Caccetta, S. Furby, and K. Zdunic, "Forest cover trends from time series Landsat data for the Australian continent," *International Journal of Applied Earth Observation and Geoinformation*, vol. 21, pp. 453–462, 2013.

[34]    C. Bayr, H. Gallaun, U. Kleb, B. Kornberger, M. Steinegger, and M. Winter, "Satellite Based Forest Monitoring: Spatial and Temporal Forecast of Growing Index and Short Wave Infrared Band," *Geospatial Health*, vol. 11(1), pp. 31–42, 2016.

[35]    R. E. Kennedy, W. B. Cohen, and T. A. Schroeder, "Trajectory-based change detection for automated characterization of forest disturbance dynamics," *Remote Sensing of Environment*, vol. 110, pp. 370 – 386, 2007.

[36] •• P. Griffiths and P. Hostert, *Remote Sensing Time Series: Revealing Land Surface Dynamics*, ch. 15: Forest Cover Dynamics During Massive Ownership Changes - Annual Disturbance Mapping Using Annual Landsat Time-Series, pp. 307–322. Springer International Publishing, 2015. **The authors showed that time series approaches can provide good results for gradual changes such as recovery or degradation in Europe, even if only annual data is available.**

[37] R. E. Kennedy, Z. Yang, and W. B. Cohen, "Detecting trends in forest disturbance and recovery using yearly Landsat time series: 1. LandTrendr - Temporal segmentation algorithms," *Remote Sensing of Environment*, vol. 114, no. 12, pp. 2897 – 2910, 2010.

[38] W. B. Cohen, Z. Yang, and R. Kennedy, "Detecting trends in forest disturbance and recovery using yearly Landsat time series: 2. TimeSync - Tools for calibration and validation," *Remote Sensing of Environment*, vol. 114, no. 12, pp. 2911–2924, 2010.

[39] N. Koutsias, M. Pleniou, G. Mallinis, F. Nioti, and N. I. Sifakis, "A rule-based semi-automatic method to map burned areas: exploring the USGS historical Landsat archives to reconstruct recent fire history," *International Journal of Remote Sensing*, vol. 34, pp. 7049–7068, oct 2013.

[40] L. Eklundh, T. Johansson, and S. Solberg, "Mapping insect defoliation in Scots pine with MODIS time-series data," *Remote Sensing of Environment*, vol. 113, pp. 1566–1573, 2009.

[41] J. Verbesselt, R. Hyndman, G. Newnham, and D. Culvenor, "Detecting trend and seasonal changes in satellite image time series," *Remote Sensing of Environment*, vol. 114, no. 1, pp. 106–115, 2010.

[42] • D. Pflugmacher, W. B. Cohen, and R. E. Kennedy, "Using landsat-derived disturbance history (1972-2010) to predict current forest structure," *Remote Sensing of Environment*, vol. 122, pp. 146–165, jul 2012. **This study demonstrates the unique value of the long, historic Landsat record, and suggests new potentials for mapping current forest structure with time series data.**

[43] K. Granica and M. Schardt, "User Utility Synthesis Report." https://www.eufodos.info/sites/default/files/reports/EF-REP-JR-2012-07-26_D510_1-synth_report_v1.pdf, 2012. Last accessed: Oct. 5ht 2016.

[44] G. P. Asner, M. Keller, R. Pereira, and J. C. Zweede, "Remote sensing of selective logging in Amazonia Assessing limitations based on detailed field observations, Landsat ETM+, and textural analysis," *Remote Sensing of Environment*, vol. 80, no. 3, pp. 483–496, 2002.

[45] G. P. Asner, D. E. Knapp, E. N. Broadbent, P. J. C. Oliveira, M. Keller, and J. N. Silva, "Selective Logging in the Brazilian Amazon," *American Association of the Advancement of Science*, vol. 310, pp. 480–482, 2005.

[46] •• M. Hansen, A. Krylov, A. Tyukavina, P. Potapov, S. Turubanova, B. Zutta, S. Ifo, B. Margono, F. Stolle, and R. Moore, "Humid tropical forest disturbance alerts using landsat data," *Environmental Research Letters*, vol. 11, no. 3, p. 034008, 2016. **This paper shows the first results of an operational forest disturbance alert system using Landsat data in three tropical countries. The results show**

**very high user's accuracies and moderately high producer's accuracies and are freely available on the internet.**

[47]     A. Koltunov, S. Ustin, G. P. Asner, and I. Fung, "Selective logging changes forest phenology in the Brazilian Amazon: Evidence from MODIS image time series analysis," *Remote Sensing of Environment*, vol. 113, pp. 2431–2440, 2009.

[48]     • M. Hirschmugl, M. Steinegger, H. Gallaun, and M. Schardt, "Mapping Forest Degradation due to Selective Logging by Means of Time Series Analysis: Case Studies in Central Africa," *Remote Sensing*, vol. 6 (1), no. ISSN 2072-4292, pp. 756–775, 2014. **Selective logging is a major driver of forest degradation in Central Africa, but often goes undetected due to the fast regrowth in tropical areas. This paper presents a method to detect the affected areas in a 10-years Landsat time series.**

[49]     P. Coppin, I. Jonckheere, K. Nackaerts, B. Muys, and E. Lambin, "Digital change detection methods in ecosystem monitoring: a review," *International Journal of Remote Sensing*, vol. 25, no. 9, pp. 1565–1596, 2004.

[50]     V. Walter, "Object-based classification of remote sensing data for change detection," *ISPRS Journal of Photogrammetry and Remote Sensing*, vol. 58, no. 3-4, pp. 225–238, 2004.

[51]     • F. Sedano, P. Kempeneers, J. S. Miguel, P. Strobl, and P. Vogt, "Towards a pan-European burnt scar mapping methodology based on single date medium resolution optical remote sensing data," *International Journal of Applied Earth Observation and Geoinformation*, vol. 20, pp. 52–59, 2013. **The authors present a two stage approach for operational burnt scar mapping with medium resolution remote sensing data in Mediterranean Europe with an increased capability for detection of smaller burnt scars.**

[52]     S. Violini, "Deforestation: Change detection in forest cover using remote sensing," in *Seminary Master in Emergency Early Warning and Response Space Applications*, (Mario Gulich Institute, CONAE. Argentina), pp. 1–28, 2013.

[53]     • A. Banskota, N. Kayastha, M. Falkowski, M. Wulder, R. Froese, and J. White, "Forest Monitoring Using Landsat Time Series Data: A Review," *Canadian Journal of Remote Sensing*, vol. 40, no. 5, pp. 362–384, 2014. **Comprehensive review of time series approaches using Landsat data including preprocessing steps and verification methods.**

[54]     C. Kuenzer, S. Dech, and W. Wagner, *Remote Sensing Time Series: Revealing Land Surface Dynamics*, ch. 1: Remote Sensing Time Series Revealing Land Surface Dynamics: Status Quo and the Pathway Ahead, pp. 1–24. Springer International Publishing, 2015.

[55]     L. Eklundh and P. Jönsson, *Remote Sensing Time Series: Revealing Land Surface Dynamics*, ch. 7: TIMESAT: A Software Package for Time-Series Processing and Assessment of Vegetation Dynamics, pp. 141–158. Springer International Publishing, 2015.

[56]     K. Gutjahr, R. Perko, H. Raggam, and M. Schardt, "The Epipolarity Constraint in Stereo-Radargrammetric DEM Generation," *Geoscience and Remote Sensing, IEEE Transactions on*, vol. Volume:52 , Issue: 8, pp. 5014 – 5022, Aug 2014.

[57]     W. Chen, W. Chen, and J. Li, "Comparison of surface reflectance derived by relative radiometric normalization versus atmospheric correction for generating large-scale landsat mosaics," *Remote Sensing Letters*, vol. 1, no. 2, pp. 103–109, 2010.

[58]     D. Schlaepfer, C. C. Borel, J. Keller, and K. I. Itten, "Atmospheric precorrected differential absorption technique to retrieve columnar water vapour," *Remote Sensing of Environment*, no. 65, pp. 353–366, 1998.

[59]     U. Mueller-Wilm, *Sentinel-2 MSI - Level-2A Prototype Processor Installation and User Manual*, 2016. Last accessed 5 Oct 2016.

[60]     O. Hagolle, M. Huc, D. Villa Pascual, and G. Dedieu, "A Multi-Temporal and Multi-Spectral Method to Estimate Aerosol Optical Thickness over Land, for the Atmospheric Correction of FormoSat-2, LandSat, VENuS and Sentinel-2 Images," *Remote Sensing*, vol. 7, no. 3, p. 2668, 2015.

[61]     C. Chance, T. Hermosilla, N. Coops, M. Wulder, and J. White, "Effect of topographic correction on forest change detection using spectral trend analysis of Landsat pixel-based composites," vol. 44, pp. 186–194, 2016.

[62]     C. Huang, N. Thomas, S. N. Goward, J. G. Masek, Z. Zhu, J. R. G. Townshend, and J. E. Vogelmann, "Automated masking of cloud and cloud shadow for forest change analysis using Landsat images," *Int. Journal of Remote Sensing*, vol. 31, pp. 5449–5464, October 2010.

[63]     Z. Zhu and C. Woodcock, "Object-based cloud and cloud shadow detection in landsat imagery," *Remote Sensing of Environment*, vol. 118, pp. 83–94, 2012.

[64]     Z. Zhu, S. Wang, and C. E. Woodcock, "Improvement and expansion of the Fmask algorithm: cloud, cloud shadow, and snow detection for Landsats 4-7, 8, and Sentinel 2 images," *Remote Sensing of Environment*, vol. 159, pp. 269 – 277, 2015.

[65]     D. C. Morton, R. S. D. Y. E. Shimabukuro, L. O. Anderson, F. D. B. Espírito-Santo, M. Hansen, and M. Carroll, "Rapid Assessment of Annual Deforestation in the Brazilian Amazon Using MODIS Data," *Earth Interactions*, vol. 9, no. 8, pp. 1–22, 2005.

[66]     C. G. Diniz, A. A. de Almeida Souza, D. C. Santos, M. C. Dias, N. C. da Luz, D. R. V. de Moraes, J. S. Maia, A. R. Gomes, I. da Silva Narvaes, D. M. Valeriano, L. E. P. Maurano, and M. Adami, "DETER-B: The New Amazon Near Real-Time Deforestation Detection System," *IEEE Journal of Selected Topics in Applied Earth Observations and Remote Sensing*, vol. 8, no. 7, 2015.

[67]     J. Miettinen, H.-J. Stibig, F. Achard, A. Langner, and S. Carboni, "Remote Sensing of Forest Degradation in Southeast Asia - Regional Review," *Asian Journal of Geoinformation*, vol. 15, pp. 23–30, 2015.

[68]     •• P. Potapov, S. Turubanova, A. Tyukavina, A. Krylov, J. McCarty, V. Radeloff, and M. Hansen, "Eastern europe's forest cover dynamics from 1985 to 2012 quantified from the full landsat archive," *Remote Sensing of Environment*, vol. 159, pp. 28–43, 2015. **The authors developed an algorithm to simultaneously process data from different Landsat platforms and sensors (TM and ETM +) to**

map annual forest cover loss and decadal forest cover gain and applied it on 59,539 Landsat images across Eastern Europe and European Russia with accuracies > 75%.

[69]    ₍₎ Z. Zhu, C. E. Woodcock, and P. Olofsson, "Continuous monitoring of forest disturbance using all available landsat imagery," *Remote Sensing of Environment*, vol. 122, pp. 75–91, 2012. **The Continuous Monitoring of Forest Disturbance Algorithm (CMFDA) presented in this paper flags forest disturbance by differencing the predicted and observed Landsat images with both producer's and user's accuracies higher than 95% in the spatial domain and temporal accuracy of approximately 94%.**

[70]    de Beurs, K. M. and G. M. Henebry, "A statistical framework for the analysis of long image time series," *International Journal of Remote Sensing*, vol. 26, no. 8, pp. 1551–1573, 2005.

[71]    M. Kuhn, "Building Predictive Models in R Using the caret Package," *Journal of Statistical Software*, vol. 28, no. 5, 2008.

[72]    A. Ghosh, F. E. Fassnacht, P. Joshi, and B. Koch, "A framework for mapping tree species combining hyperspectral and LiDAR data: Role of selected classifiers and sensor across three spatial scales," *International Journal of Applied Earth Observation and Geoinformation*, vol. 26, pp. 49–63, Feb. 2014.